# Victor Davis

## Types, Tokens, and Hapaxes: A New Heap's Law

### Abstract

Heap's Law[1] states that in a large enough text corpus, the number of types as a function of tokens grows as $N = KM^\beta$ for some free parameters $K, \beta$. Much has been written[2,3,4,5,6] about how this result and various generalizations can be derived from Zipf's Law[7]. Here we derive from first principles a completely novel expression of the type-token curve and prove its superior accuracy on real text. This expression naturally generalizes to equally accurate estimates for counting hapaxes and higher $n$-legomena.

### Introduction

Zipf's Law is usually formulated[7] as Freq $\propto$ 1/Rank, generalized by raising Rank to an exponent $\alpha$. This is benchmarked against real text and shown to be curiously accurate for corpora of varying sizes but exhibiting a fat tail. Therefore it performs most poorly on rare words, which contribute the most to the type-token curve. The common interpretation is that any inaccuracy in modeling a type-token growth curve can be explained away by this fat tail in which the unpredictability of the frequency of rare words introduces noise into an otherwise correct signal. This paradigm is completely false. We show that Zipf's Law can be reformulated to model the frequencies of rare words as accurately as common ones and that this reformulation leads to a simple, intuitive derivation of a logarithmic (*not* exponential) type-token growth curve.

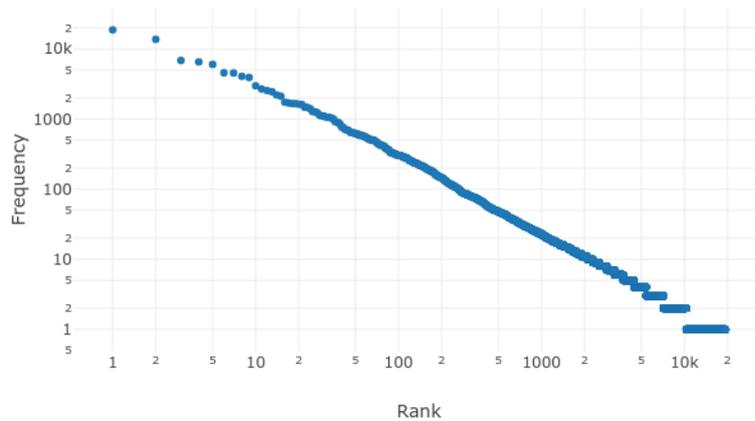

## Keywords

**Tokens**: Instances of words in a text, elements of the corpus in which words are considered an ordered list. (Using "words" and "tokens" interchangeably.)

**Types**: Distinct words of a text, elements of the lexicon/vocabulary or the size of the *set* of tokens. (Using "distinct words" and "types" interchangeably.)

**Hapaxes**: Types that occur exactly once in a corpus.

**Dis, Tris, Tetrakis, Pentakis, $n$-Legomena**: Types that occur exactly $n$ times in a corpus.

**$(M, N)$-corpus**: A text corpus consisting of $M$ tokens and $N$ types.

**Optimum Sample**: A text corpus exhibiting a perfect Zipf distribution.

**Perfect Zipf Distribution**: A word frequency distribution following Zipf's law literally and perfectly, reformulated to directly estimate the number of rare words. (Defined in more detail later.)

## Hypothesis

For a text corpus consisting of $M$ tokens and $N$ types, randomly sampling $m$ tokens of that corpus yields $y$ types, and the expected value of $y \approx E(m)$ is given by (1), parametrized by $(M_z, N_z)$, the size of an optimum sample.

$$E(m) = N_z \ln\left(\frac{m}{M_z}\right) \frac{\frac{m}{M_z}}{\frac{m}{M_z} - 1} \qquad (1)$$

The procedure for obtaining the parameters $(M_z, N_z)$ is described in the "Results" section later, equations (19) and (20), respectively.

## Methods

For some small $n$, construct a $(kn, k)$-corpus by collecting every word that occurs *exactly* $n$ times. This mini-corpus, a sub-selection of the whole, has a perfectly uniform word frequency distribution, a text consisting of $kn$ total words (tokens): $k$ distinct words (types) each repeating exactly $n$ times. What is $E_n(m)$, the expected number of types in a random selection of $m$ words out of this mini-corpus? If sampling sequentially without replacement, the expected value is the partial sum of the probability at each step of drawing a new type.

$$E_n(m+1) = E_n(m) + P_{new}(m+1) \qquad (2)$$

Suppose $m$ tokens are drawn at random without replacement, resulting in $y$ types. There are then $kn - m$ tokens left to draw, $n(k - y)$ of which are of a type not yet drawn. Thus,

$$P_{new}(m+1) = \frac{n(k-y)}{kn-m} \qquad (3)$$

$$E_n(m+1) = y + \frac{n(k-y)}{kn-m} \qquad (3a)$$

Substituting the *actual* number of types drawn, $y$, for the *expected* number of types drawn, $E_n(m)$, we have a recursive expression for the growth of the curve $(m, y)$ starting at $E_n(0) = 0$.

$$E_n(m+1) = E_n(m) + \frac{n(k - E_n(m))}{kn - m} \qquad (4)$$

This recursion performs quite well with real data. Consider a shuffled deck of cards, analogous to either a $(52, 4)$- or $(52, 13)$-corpus, depending whether suits or valors are considered "types."

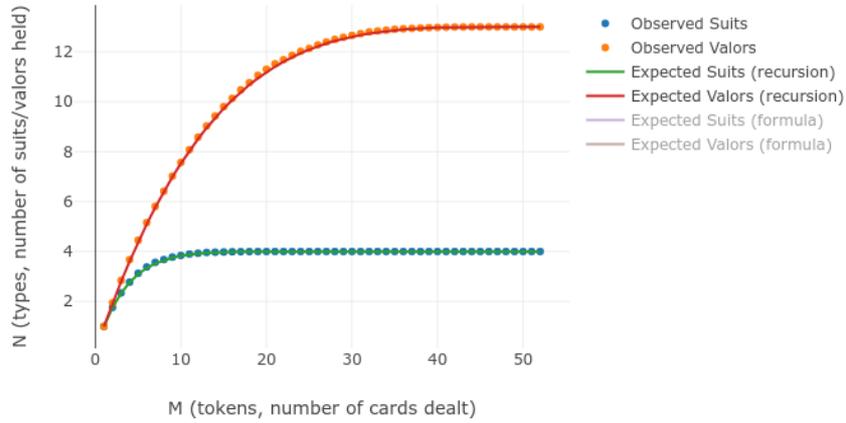

We can use calculus to derive a non-recursive version of this function. Since the derivative is approximately equal to the vertical distance between successive elements in the series, we can rewrite (4) and integrate.

$$\frac{dE}{dm} \approx \frac{E_n(m+1) - E_n(m)}{m+1-m} = P_{new}(m+1) = \frac{n(k-E)}{kn-m} \quad (5)$$

$$\int \frac{dE}{k-E} \approx n \int \frac{dm}{kn-m} \rightarrow -ln(k-E) \approx -n\ln(kn-m) + C$$

$$E_n(m) \approx k - C'(kn-m)^n \quad (6)$$

Using $E_n(0) = 0$ as a boundary condition, we find $C' = \frac{k}{(kn)^n}$, giving:

$$E_n(m) \approx k - k\left(1 - \frac{m}{kn}\right)^n \quad (7)$$

$$Suits(m) = E_{13}(m) \approx 4 - 4\left(1 - \frac{m}{52}\right)^{13} \quad (7a)$$

$$Valors(m) = E_4(m) \approx 13 - 13\left(1 - \frac{m}{52}\right)^4 \quad (7b)$$

Again, we find a near-perfect match between the discrete values calculated by recursing and the continuous values given by (7). *Near* perfect because the discrete case (actual reality) effectively stopped the integral limit approaching zero once $dm = 1$. Thus, (7) is an analytical estimate of (4) smoothing out the discontinuities found in a real deck of cards.

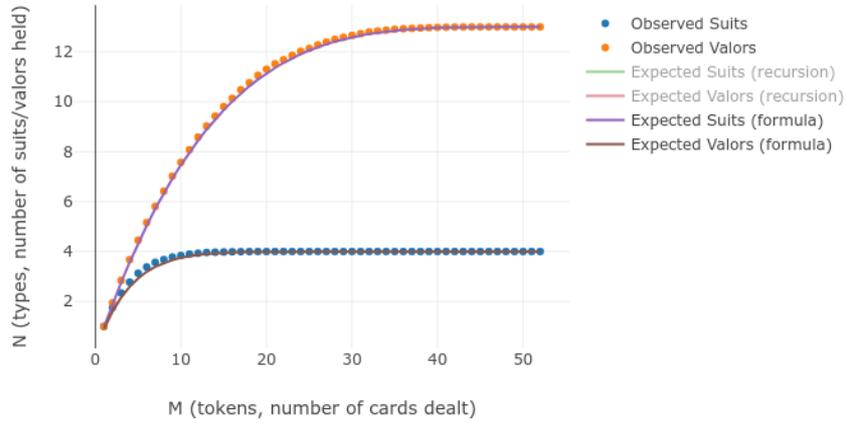

## Text Corpora

A $(M, N)$-corpus can be partitioned into a number of $(kn, k)$-mini-corpuses, and the expected type-token curve of the whole can be gotten by summing over the individual parts. Start by partitioning the corpus into decks: all hapaxes into the first deck, all dis legomena into the second deck, tris into the third deck, etc. The $n$th deck shall be a $(nk_n, k_n)$-mini-corpus, the whole corpus encoded by some vector $\hat{k} = (k_0, k_1, k_2, k_3, \ldots)$. (Call the "zeroth deck" the null set for now, for $k_0 = 0$ number of types occurring exactly zero times in the whole corpus, which will come in handy shortly.)

The whole corpus consists of all these decks randomly shuffled together. Sampling $m$ tokens should yield on average $\frac{nk_n}{M}m$ tokens from the $n$th deck. The number of types yielded from each deck is therefore $E_n\left(\frac{nk_n}{M}m\right)$. Substitute this into (7) and sum over all decks.

$$E(m) = \sum E_n\left(\frac{nk_n}{M}m\right) = \sum k_n - k_n\left(1 - \frac{m}{M}\right)^n = N - \sum k_n\left(1 - \frac{m}{M}\right)^n$$

As for the bounds on the summation, we can consider $\hat{k}$ a finite vector in infinite-dimensional space with zeroes imputed into missing indexes (numbers $n$ for which no words occur *exactly* $n$ times). This expression (8), as we'll see later in the "Results" section, performs nearly perfectly on real data. We can also rewrite it as an infinite series over the interval $[0, 1]$ by substituting $x = \frac{m}{M}$ the proportion of the corpus sampled rather than the raw token count.

$$E(m) = N - \sum_{n=0}^{\infty} k_n\left(1 - \frac{m}{M}\right)^n \qquad (8)$$

$$E(x) = N - \sum_{n=0}^{\infty} k_n (1-x)^n \qquad (8b)$$

What does (8b) really *mean*? What happens when we sample $m$ tokens at random, or some proportion $x$ of the corpus? Choosing a sample size of $x = 1$ gives a permutation of the original corpus, and it's easy to see from (8b) that $E(1) = N$ as expected, i.e. all types are drawn. Sampling some proportion $x$ of the corpus, we expect $E(x)$ types to be drawn, and $N - E(x)$ types *not* to be drawn. What is the expected value of the latter? For the hapaxes, the probability of *not* drawing each token is $1 - x$, so the expected number of hapaxes *not* drawn is $k_1(1-x)$. The probability of *not* drawing *both* instances of each dis legomenon is $(1-x)^2$ so the expected number of dis legomena *not* drawn is $k_2(1-x)^2$. Continuing this argument, we define $k_0(x)$, the expected number of types *not* drawn when sampling proportion $x$ of the corpus:

$$k_0(x) = k_0 + k_1(1-x) + k_2(1-x)^2 + k_3(1-x)^3 + \ldots = N - E(x) \quad (9)$$

It can be seen that (9) is just a way of rewriting (8b) given this new intuition. But let's go further. How many *hapaxes* can be expected when sampling proportion $x$ of the corpus? The expected number of existing hapaxes drawn is $k_1 x$. But a hapax will be *created* in the sample if *one* instance of a dis legomenon is drawn but not the other. We should expect $\binom{2}{1} k_2 x(1-x)$ to be created this way. Likewise, we expect to create a hapax in our sample by drawing only *one* of the three instances of each tris legomenon, resulting in $\binom{3}{1} k_3 x(1-x)^2$ created this way. In general, using the shorthand $k_n = k_n(1)$,

$$k_1(x) = k_1 x + 2k_2 x(1-x) + 3k_3 x(1-x)^2 + 4k_4 x(1-x)^3 + \ldots \quad (10)$$

$$k_n(x) = \sum_{i=n}^{\infty} \binom{i}{n} k_i x^n (1-x)^{i-n} \qquad (11)$$

There is a beautiful way of visualizing this as a matrix transformation acting on $\hat{k}$ involving Pascal's Triangle. Since each of the columns sum to $1$, it's easy to prove that both the input vector $\hat{k}$ and the resultant vector $\hat{k}(x)$ sum to $N$. The equation (11) above expresses the row sums.

$$\begin{bmatrix} k_0(x) \\ k_1(x) \\ k_2(x) \\ k_3(x) \\ k_4(x) \\ \vdots \end{bmatrix} = \begin{bmatrix} 1 & (1-x) & (1-x)^2 & (1-x)^3 & (1-x)^4 & \cdots \\ 0 & x & 2x(1-x) & 3x(1-x)^2 & 4x(1-x)^3 & \cdots \\ 0 & 0 & x^2 & 3x^2(1-x) & 6x^2(1-x)^2 & \cdots \\ 0 & 0 & 0 & x^3 & 4x^3(1-x) & \cdots \\ 0 & 0 & 0 & 0 & x^4 & \cdots \\ \vdots & \vdots & \vdots & \vdots & \vdots & \ddots \end{bmatrix} \begin{bmatrix} k_0 \\ k_1 \\ k_2 \\ k_3 \\ k_4 \\ \vdots \end{bmatrix}$$

$$\hat{k}(x) = A_x \hat{k} \qquad (12)$$

$$\hat{k}(0) = A_0 \hat{k} = (N, 0, 0, 0, \ldots) \qquad (12a)$$

$$\hat{k}(1) = A_1 \hat{k} = I\hat{k} = (0, k_1, k_2, k_3, \ldots) \quad (12b)$$

(If the idea of infinite matrices doesn't sit well, remember $\hat{k}$ is always finite in practice.) Notice (11) can be expressed in terms of the $n$th derivatives of $E$ for $n > 0$:

$$k_n(x) = (-1)^n \frac{x^n}{n!} k_0^{(n)}(x) = (-1)^{n+1} \frac{x^n}{n!} E^{(n)}(x) \quad (14)$$

By now we've achieved three results, having not yet made a single empirical assumption about the word frequency distribution of a corpus. Any type-token system describable in terms of the evolution of some $\hat{k}$ follows these rules.

1. Equation (12) gives an algorithmic approach to predicting $n$-legomena counts when sampling a corpus. Using the observed counts from the *whole* corpus as input, the word frequency distribution of any sample can be simulated by constructing $A_x$ and transforming $\hat{k}$.
2. Any function such as Heap's Law which approximates the type-token growth curve can be generalized via (14) to make testable predictions for $n$-legomena counts as well.
3. If $k_n$ can be approximated by a series and $E(x)$ converges on some analytic function, then (14) gives analytic functions for $n$-legomena as well.

Can $\hat{k}$ be approximated by a series? In real text, do the tabulations of $n$-legomena actually follow some regular, enumerative pattern?

## Zipf's Law

Suppose that for a given $(M, N)$-corpus there exists a corresponding optimum sample $(M_z, N_z)$-corpus following a *perfect Zipf distribution*. If the original corpus is "too big" then some number of randomly selected tokens $M_z < M$ will produce a sub-selection of text exhibiting Zipf's Law. If the original corpus is "too small" then it can be thought of as a sub-selection of some larger corpus consisting of $M_z > M$ tokens exhibiting Zipf's Law. (Of course, some corpora may be "just right" and $M_z = M$.)

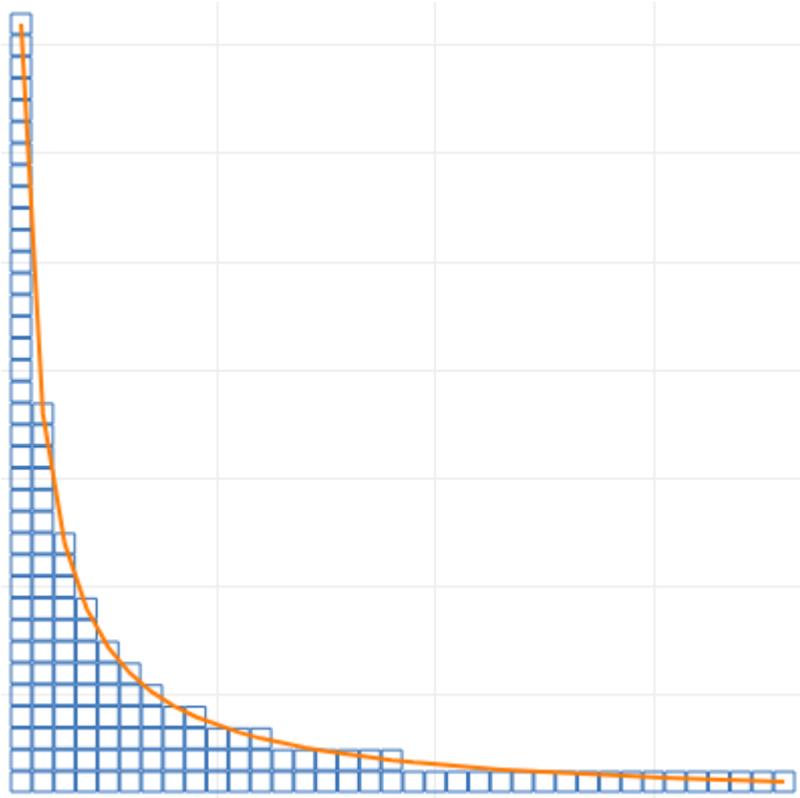

Define a *perfect Zipf distribution* as follows: For some $(M_z, N_z)$-corpus with a ranked word frequency distribution $(f_1, f_2, f_3, \dots)$, the expected number of types repeating $n$ or more times $\pi(f_r \geq n)$ is roughly equal to $\frac{1}{n} N_z$. Alternatively, there exists about $\frac{1}{n} N_z$ types repeating $n$ or more times. "Roughly" and "about" to allow for normal statistical noise in the still "perfect" sample.

**Corollaries:**

- $k_{a<b}$ the expected number of types repeating between $a$ and $b$ times is $\pi(f_r \geq a) - \pi(f_r \geq b) = \left(\frac{1}{a} - \frac{1}{b}\right) N_z = \frac{b-a}{ab} N_z$
- The expected number of types repeating fewer than $n$ times is $\pi(f_r < n) = \frac{n-1}{n} N_z$
- No types occur exactly zero times: $k_0 = \pi(f_r < 1) = \frac{1-1}{1} N_z = 0$
- $k_n$ the expected number of types repeating *exactly* $n$ times, or the expected number of $n$-legomena (for $n > 0$) equals
  $\pi(f_r \geq n) - \pi(f_r \geq n+1) = \left(\frac{1}{n} - \frac{1}{n+1}\right) N_z = \frac{1}{n(n+1)} N_z$
- The expected *proportions* of $n$-legomena are
  $\left\{\frac{1}{2}, \frac{1}{6}, \frac{1}{12}, \frac{1}{20}, \frac{1}{30}, \dots\right\}$
- This distribution is already normalized: $\sum_{n=1}^{\infty} \frac{N_z}{n(n+1)} = N_z$
- The expected frequency of the most common word is $f_1 = \frac{1}{1} N_z = N_z$
- The expected frequency of the $r$th most common word is
  $f_r = \left\lfloor \frac{N_z}{r} \right\rfloor$
- Freq $\propto$ 1/Rank

This reformulation of Zipf's Law is practically equivalent to the original, with the caveats that (a) for high ranks (rare words) we round *down* to the nearest integer to obtain the expected frequency, and (b) the total number of types is equal to the frequency of the commonest word. (So, curiously, there is one "the" for each vocabulary word in the lexicon.) This leaves us with a specific prediction for the counts of rare words ($n$-legomena for low $n$). The expected number of hapaxes, dis, tris, and higher $n$-legomena, for $n > 0$ is given by:

$$k_n = \frac{N_z}{n(n+1)} \tag{15}$$

Substituting (15) into (8b) above, and taking $k_0 = 0$,

$$E(x) = N_z - N_z \sum_{n=1}^{\infty} \frac{(1-x)^n}{n(n+1)} \tag{16}$$

It can be shown that (16) converges on (17). Taking successive derivatives, per (14), we now have expressions not just for the number of types with respect to tokens, but also hapaxes and higher $n$-legomena:

$$E(x) = N_z \frac{\ln(x)x}{x-1} \tag{17}$$

$$k_0(x) = N_z - E(x) = N_z \frac{x - \ln(x)x - 1}{x - 1} \tag{17.0}$$

$$k_1(x) = xE'(x) = N_z \frac{x^2 - \ln(x)x - x}{(x-1)^2} \tag{17.1}$$

$$k_2(x) = -\frac{x^2}{2} E''(x) = N_z \frac{x^3 - 2\ln(x)x^2 - x}{2(x-1)^3} \tag{17.2}$$

$$k_3(x) = \frac{x^3}{6} E'''(x) = N_z \frac{2x^4 + 3x^3 - 6\ln(x)x^3 - 6x^2 + x}{6(x-1)^4} \tag{17.3}$$

$$k_4(x) = N_z \frac{3x^5 + 10x^4 - 12x^4 \ln(x) - 18x^3 + 6x^2 - x}{12(x-1)^5} \tag{17.4}$$

$$k_5(x) = N_z \frac{12x^6 + 65x^5 - 60x^5 \ln(x) - 120x^4 + 60x^3 - 20x^2 + 3x}{60(x-1)^6} \tag{17.5}$$

## Results

The King James Bible is a $(1010654, 13769)$-corpus. The most common word "the" appears $62103$ times (not $13769$), and $h_{obs} = 32.06\%$ of the types are hapaxes (not $50\%$). It is nowhere near an optimum sample, so how do we find a suitable $(M_z, N_z)$? The proportion of hapaxes is a decreasing function, starting at $H(0) = 1$, falling to $H(1) = \frac{1}{2}$, and continuing to fall as $x$ goes to

infinity. We can model this by expressing (17.1) as a proportion of (17) to obtain:

$$H(x) = \frac{k_1(x)}{E(x)} = \frac{1}{\ln(x)} + \frac{1}{1-x} \qquad (18)$$

In the limit, $H(0) = 1, H(1) = \frac{1}{2}$ as expected, even though the function is undefined at those points. Using a binary search algorithm, we find that $H(10.41) = .3206$. Could this mean the Bible is ten times "too large" to exhibit Zipf's Law? Taking $M_z = M/10.41$ and scaling $N_z$ such that $E(M) = N$, we fit (1) to our sample corpus using optimum sample parameters $(M_z, N_z) = (97084, 5312)$. Calculating $z = H^{-1}(h_{obs})$, we have:

$$M_z = M/z \qquad (19)$$

$$N_z = N\frac{(z-1)}{\ln(z)z} \qquad (20)$$

$$E(m) = 5312 \ln(m/97084)\frac{m/97084}{m/97084 - 1} \qquad (1.1)$$

**Observed vs Predicted: King James Bible**

| TOKENS | TYPES | HAPAX | TYPES PRED | HAPAX PRED |
|---|---|---|---|---|
| 4,042 | 993 | 632 | 734 | 535 |
| 105,092 | 5,766 | 2,482 | 5,525 | 2,726 |
| 206,142 | 7,781 | 3,124 | 7,561 | 3,310 |
| 307,192 | 9,075 | 3,531 | 8,946 | 3,633 |
| 408,242 | 10,087 | 3,750 | 10,010 | 3,846 |
| 509,292 | 10,898 | 3,881 | 10,878 | 4,001 |
| 610,342 | 11,664 | 4,050 | 11,613 | 4,120 |
| 711,392 | 12,303 | 4,213 | 12,252 | 4,215 |
| 812,442 | 12,855 | 4,283 | 12,817 | 4,294 |
| 913,492 | 13,314 | 4,353 | 13,324 | 4,359 |
| 1,010,654 | 13,769 | 4,414 | 13,767 | 4,413 |

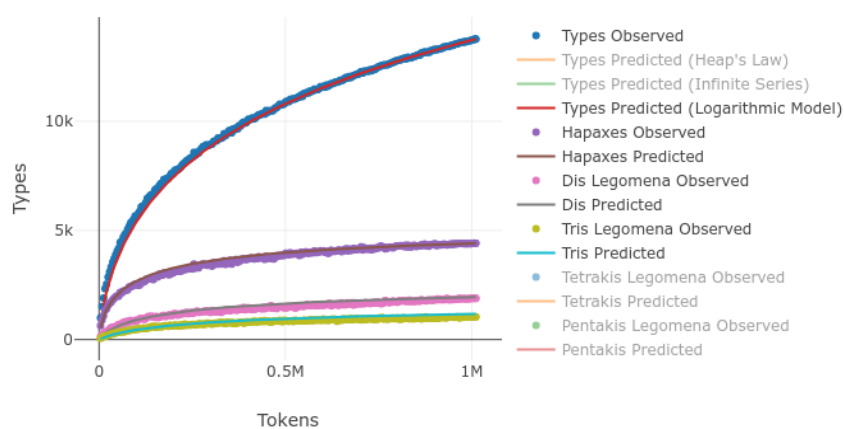

William Blake's *Poems* is a $(8354, 1820)$-corpus. The most common word "the" appears $351$ times (not $1820$) and $h_{obs} = 55.44$% of its types are hapaxes. Using a binary search, we find that $H(.5182) = .5544$, making *Poems* "too small" by half. Using the same procedure as above, we find $(M_z, N_z) = (16121, 2574)$.

$$E(m) = 2574 \ln(m/16121) \frac{m/16121}{m/16121 - 1} \quad (1.2)$$

### Observed vs Predicted: William Blake's Poems

| TOKENS | TYPES | HAPAX | TYPES PRED | HAPAX PRED |
|---|---|---|---|---|
| 33 | 25 | 23 | 33 | 27 |
| 858 | 439 | 324 | 424 | 304 |
| 1,683 | 704 | 491 | 678 | 457 |
| 2,508 | 909 | 587 | 882 | 571 |
| 3,333 | 1,066 | 653 | 1,057 | 662 |
| 4,158 | 1,211 | 740 | 1,212 | 739 |
| 4,983 | 1,366 | 809 | 1,352 | 805 |
| 5,808 | 1,495 | 878 | 1,480 | 864 |
| 6,633 | 1,604 | 931 | 1,598 | 916 |
| 7,458 | 1,707 | 953 | 1,708 | 963 |
| 8,283 | 1,812 | 1,010 | 1,811 | 1,005 |

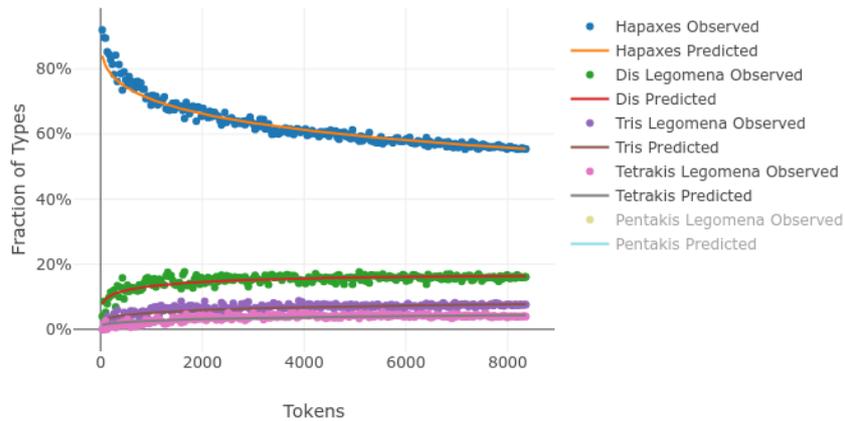

Repeating the procedure, we obtain values for $(M_z, N_z)$ that estimate types, hapaxes, and higher $n$-legomena for a number of different books. We'll evaluate the model fit using root mean square error as a percent of observed types. Below are the fitting errors for real data. The fit for (8) using actual $n$-legomena counts as the coefficients gives a good baseline for measuring the model, since it doesn't assume Zipf's Law. Making the leap from (8) to (17) requires the assumption of the existence of an optimum sample following a perfect Zipf distribution, an assumption that appears to be more accurate in some cases, like Melville's *Moby Dick*, and less accurate in others, like Shakespeare's *Julius Caesar*. Still, Heap's Law is consistently accurate only to within $2 - 3$%, while the logarithmic function derived above is nearly always accurate to within $1$%.

## RMSE % Comparison for Various Books

| TITLE | $M_z$ | $N_z$ | HEAPS | SERIES | MODEL |
|---|---|---|---|---|---|
| Moby Dick by Herman Melville 1851 | 173,272 | 15,854 | 2.66% | 0.23% | 0.23% |
| Leaves of Grass by Walt Whitman 1855 | 144,542 | 13,845 | 2.50% | 0.29% | 0.35% |
| The Ball and The Cross by G. K. Chesterton 1909 | 91,990 | 8,714 | 2.27% | 0.36% | 0.39% |
| The Wisdom of Father Brown by G. K. Chesterton 1914 | 79,540 | 7,980 | 2.44% | 0.41% | 0.43% |
| Paradise Lost by John Milton 1667 | 113,779 | 11,666 | 2.23% | 0.31% | 0.47% |
| The Man Who Was Thursday by G. K. Chesterton 1908 | 69,764 | 6,807 | 2.17% | 0.40% | 0.59% |
| Stories to Tell to Children by Sara Cone Bryant 1918 | 26,223 | 3,108 | 2.56% | 0.51% | 0.68% |

| TITLE | $M_z$ | $N_z$ | HEAPS | SERIES | MODEL |
|---|---|---|---|---|---|
| Poems by William Blake 1789 | 16,121 | 2,574 | 2.69% | 0.72% | 0.74% |
| Sense and Sensibility by Jane Austen 1811 | 38,414 | 3,817 | 2.63% | 0.38% | 0.78% |
| Alice ' s Adventures in Wonderland by Lewis Carroll 1865 | 18,864 | 2,275 | 2.44% | 0.60% | 0.79% |
| The Parent ' s Assistant , by Maria Edgeworth | 67,166 | 5,716 | 2.43% | 0.34% | 0.80% |
| The Tragedie of Hamlet by William Shakespeare 1599 | 182,050 | 13,320 | 1.07% | 0.42% | 0.97% |
| The Adventures of Buster Bear by Thornton W . Burgess 1920 | 7,158 | 1,127 | 2.75% | 0.71% | 0.99% |
| Persuasion by Jane Austen 1818 | 48,342 | 4,393 | 2.23% | 0.40% | 1.00% |
| The Tragedie of Macbeth by William Shakespeare 1603 | 106,245 | 9,465 | 1.50% | 0.52% | 1.02% |
| Emma by Jane Austen 1816 | 57,337 | 4,528 | 2.30% | 0.36% | 1.07% |
| The Tragedie of Julius Caesar by William Shakespeare 1599 | 64,786 | 5,838 | 1.51% | 0.54% | 1.09% |
| The King James Bible | 97,084 | 5,312 | 2.25% | 0.28% | 1.13% |

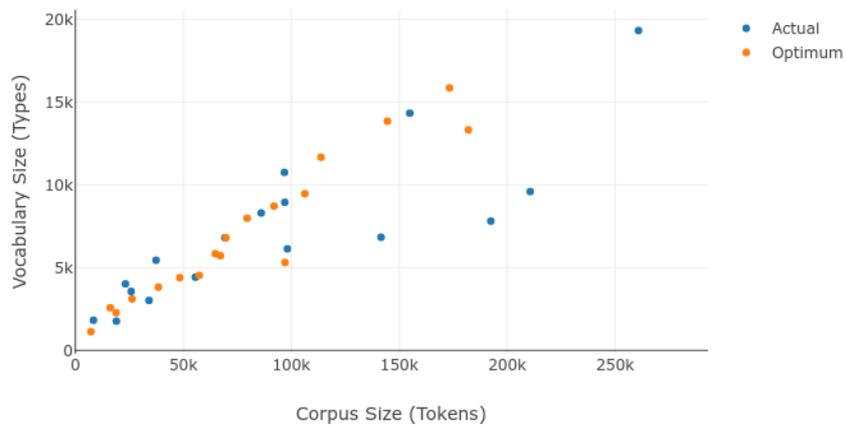

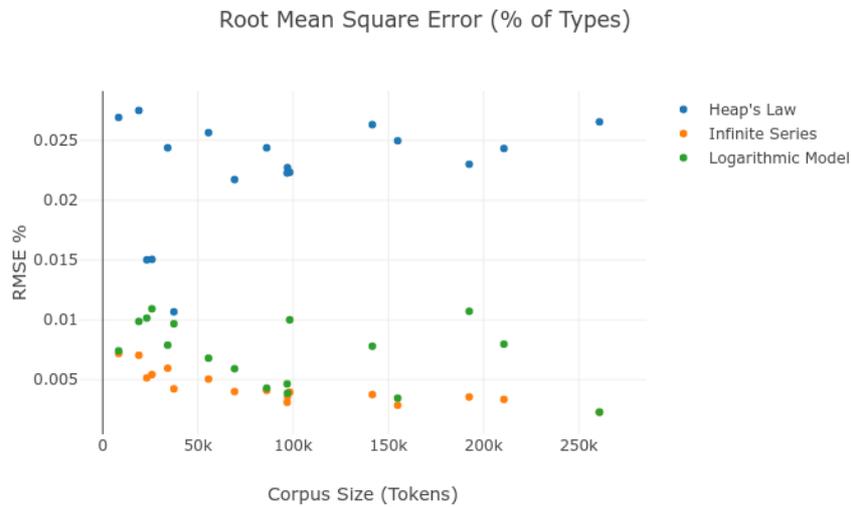

## Discussion

It's worth reinforcing that Zipf's Law is only accurate when applied to a certain optimum sample size of a given corpus. Smaller samples *won't* follow Zipf's Law, but as text accumulates, the sample will reach a phase at which it exhibits Zipf's Law peculiarly well. As text continues to accumulate, the pattern deteriorates in the opposite direction. The logarithmic functions (17) accurately model this behavior even if they can't explain the motivation for it. But it has to be stressed that Zipf's Law is *not* a pattern accurately describing all texts; it is better thought of as describing a phase text passes through as it accumulates.

This phase is the point at which (a) there are approximately as many instances of the most common word (usually "the") as there are distinct words in the vocabulary, and (b) the proportions of hapaxes, dis, tris, etc legomena with respect to the number of types is roughly $\left\{\frac{1}{2}, \frac{1}{6}, \frac{1}{12}, \frac{1}{20}, \frac{1}{30}, \dots\right\}$. It should not be expected that (a) and (b) *always* hold, and that some texts follow this more accurately than others. Our hypothesis is that it seems to be the case that *all* texts follow (a) and (b) peculiarly well at some optimum sample size, and that by discovering this optimum, accurate predictions can be made for how these proportions evolve as the sample size is varied.

## Conclusion

Most investigations of Zipf's and Heap's respective Laws and their inter-relationship tolerate a conspicuous degree of error when applied to real text. This may reflect an implicit bias that these "laws" are mere curiosities of nature, that no model *should* be able to fit cogent, rational human language to a tidy physics equation. Shockingly, the discovery of the formulas above and their accurate fit to real data imply that, far from being mere curiosities, Zipf's Law and its corresponding derivations for estimating types, hapaxes, and higher $n$-legomena must be more fundamental to the nature of language than we give it credit. It's still an open question *why* Zipf's Law seems to hold. Taking it in a very literal formulation as an axiom, accurate predictions can be made about real text, bolstering the idea that *something* fundamental must be going on, and Zipf's original observation must be the inevitable cumulative effect of some deeper, underlying information-theoretic property of human language.

## Correspondence Address

Victor Davis victor.davis.us@member.mensa.org

## References

| 1 | 2 | 3 | 4 | 5 | 6 | 7 | 8 | 9 | 10 | 11 | 12 | 13 | 14 |

1. https://dl.acm.org/citation.cfm?id=539986 Heaps, H S 1978 *Information Retrieval: Computational and Theoretical Aspects* (Academic Press)↵↵

2. http://iopscience.iop.org/article/10.1088/1367-2630/15/9/093033 Font-Clos, Francesc 2013 *A scaling law beyond Zipf's law and its relation to Heaps' law* (New Journal of Physics 15 093033)↵↵

3. http://iopscience.iop.org/article/10.1088/1367-2630/11/12/123015 Bernhardsson S, da Rocha L E C and Minnhagen P 2009 *The meta book and size-dependent properties of written language* (New Journal of Physics 11 123015)↵↵

4. http://iopscience.iop.org/article/10.1088/1742-5468/2011/07/P07013 Bernhardsson S, Ki Baek and Minnhagen 2011 *A paradoxical property of the monkey book* (Journal of Statistical Mechanics: Theory and Experiment, Volume 2011)↵↵

5. http://milicka.cz/kestazeni/type-token_relation.pdf Milička, Jiří 2009 *Type-token & Hapax-token Relation: A Combinatorial Model* (Glottotheory. International Journal of Theoretical Linguistics 2 (1), 99–110)↵↵

6. https://www.nature.com/articles/srep00943 Petersen, Alexander 2012 *Languages cool as they expand: Allometric scaling and the decreasing need for new words* (Scientific Reports volume 2, Article number: 943)↵↵

7. http://dx.doi.org/10.1037/h0052442 Zipf, George 1949 *Human behavior and the principle of least effort* (Reading: Addison-Wesley)↵↵↵

8. www.hup.harvard.edu/catalog.php?isbn=9780674434929 Zipf, George 1932 *Selective studies and the principle of relative frequency in language* (Cambridge: Harvard University Press)↵

9. https://catalog.hathitrust.org/Record/000359461 Zipf, George 1935 *The psycho-biology of language: An introduction to dynamic philology* (Boston: Mifflin Company)↵

10. https://www.journals.uchicago.edu/doi/abs/10.1086/364570 Herdan, Gustav 1960 *Type-token mathematics* (The Hague: Mouton)↵

11. https://dl.acm.org/citation.cfm?id=325476 Baeza-Yates, Ricardo 1997 *Block Addressing Indices for Approximate Text Retrieval* (Journal of the American Society for Information Science, v.51 n.1, p.69-82, Jan. 2000)↵

12. http://dx.doi.org/10.1155/2012/480196 Chen, Yanguang 2012 Zipf's law, *Hierarchical Structure, and Shuffling-Cards Model for Urban Development* (Discrete Dynamics in Nature and Society, Volume 2012)↵

13. http://iopscience.iop.org/article/10.1088/1367-2630/13/4/043004 Seung, Ki Baek 2011 *Zipf's law unzipped* (New Journal of Physics 13 043004)↵

14. https://doi.org/10.1016/j.physa.2011.05.003 Eliazar, Iddo 2011 *The growth statistics of Zipfian ensembles: Beyond Heaps' law* (Physica A, Volume 390, Issue 20)↵